\documentclass[oribibl]{llncs}

\usepackage{type1cm}        % activate if the above 3 fonts are
\usepackage{multirow}
\usepackage{caption}
\usepackage{graphicx}        % standard LaTeX graphics tool
\usepackage{algorithm}
\usepackage{algorithmic}
\usepackage{fixltx2e}
\usepackage[section]{placeins}
\usepackage{float}
\usepackage[utf8]{inputenc}

\usepackage[usestackEOL]{stackengine}

\usepackage{amsmath}
\usepackage{array}
\usepackage[final]{pdfpages}
\usepackage[ampersand]{easylist}
\usepackage{color}

\usepackage{array}
\newcolumntype{P}[1]{>{\centering\arraybackslash}p{#1}}

\begin{document}

\frontmatter
\mainmatter

\title{Method for aspect-based sentiment annotation using rhetorical analysis}

\author{Łukasz Augustyniak\inst{1} \and Krzysztof Rajda\inst{2} \and Tomasz Kajdanowicz\inst{1}}
% \authorrunning{} 

\institute{Wroclaw University of Technology, Wroclaw, Poland \\
Department of Computational Intelligence, Wroclaw University of Technology 
\email{lukasz.augustyniak@pwr.edu.pl, tomasz.kajdanowicz@pwr.edu.pl}
\and
Kenaz Technologies
\email{kenaz.technologies@gmail.com}
}

\maketitle

\begin{abstract}
This paper fills a~gap in aspect-based sentiment analysis and aims to present a~new method for preparing and analysing texts concerning opinion and generating user-friendly descriptive reports in natural language. We present a~comprehensive set of techniques derived from Rhetorical Structure Theory and sentiment analysis to extract aspects from textual opinions and then build an abstractive summary of a set of opinions. Moreover, we propose aspect-aspect graphs to evaluate the importance of aspects and to filter out unimportant ones from the summary. Additionally, the paper presents a prototype solution of data flow  with interesting and valuable results. The proposed method's results proved the high accuracy of aspect detection when applied to the gold standard dataset. 

\end{abstract}

\section{Introduction}
\label{sec:Introduction}

Modern society is an information society bombarded from all sides by an increasing number of different pieces of information. The 21st century has brought us the rapid development of media, especially in the internet ecosystem. This change has caused the transfer of many areas of our lives to virtual reality. New forms of communication have been established. Their development has created the need for analysis of related data. Nowadays, unstructured information is available in digital form, but how can we analyse and summarise billions of newly created texts that appear daily on the internet? Natural language analysis techniques, statistics and machine learning have emerged as tools to help us. In recent years, particular attention has focused on  sentiment analysis. This area is defined as the study of opinions expressed by people as well as attitudes and emotions about a~particular topic, product, event, or person. Sentiment analysis determines the polarisation of the text. It answers the question as to whether a~particular text is a~positive, negative, or neutral one.

\begin{figure}[!ht]
\centering
\includegraphics[scale=0.1]{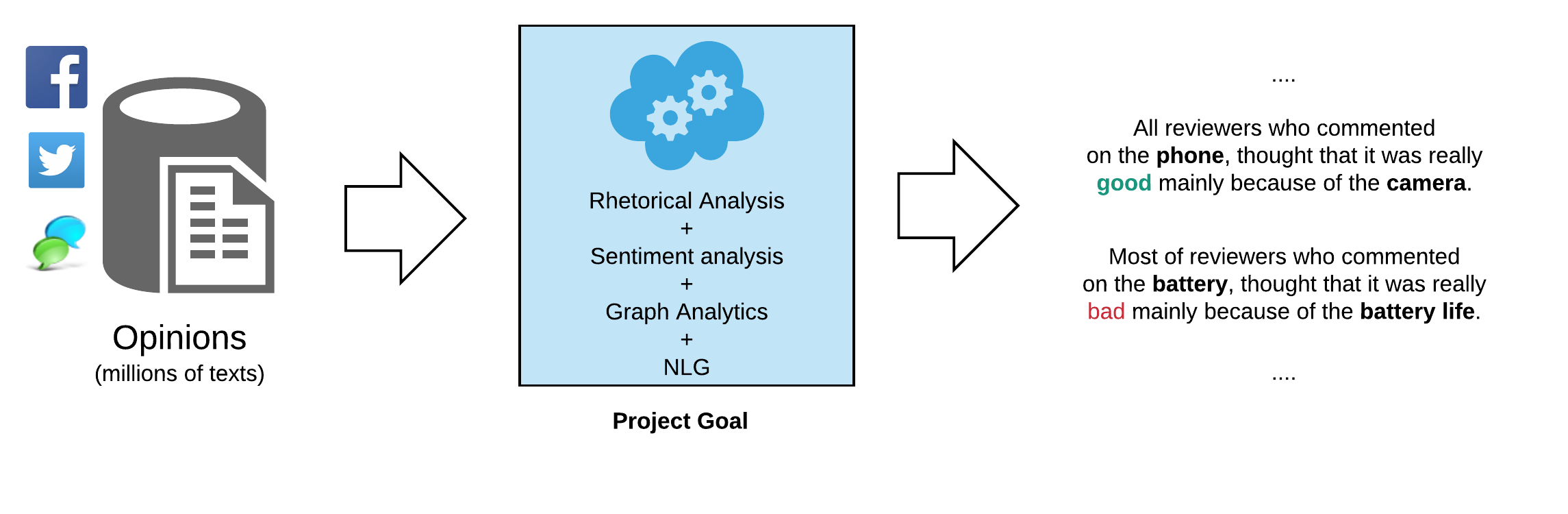}
\caption{The workflow for Rhetorical and Sentiment Analysis.\label{fig:introflow}}
\end{figure}

Our goal is to build a~comprehensive set of techniques for preparing and analysing texts containing opinions and generating user-friendly descriptive reports in natural language - Figure \ref{fig:introflow}. In this paper, we describe briefly the whole workflow and present a prototype implementation. Currently, existing solutions for sentiment annotation offer mostly analysis on the level of entire documents, and if you go deeper to the level of individual product features, they are only superficial and poorly prepared for the analysis of large volumes of data. This can especially be seen in scientific articles where the analysis is carried out on a~few hundred reviews only. It is worth mentioning that this task is extremely problematic because of the huge diversity of languages and the difficulty of building a~single solution that can cover all the languages used in the world. Natural language analysis often requires additional pre-processing steps, especially at the stage of preparing the data for analysis, and steps specific for each language. Large differences can be seen in the analysis of the Polish language (a highly inflected language) and English (a grammatically simpler one). We propose a~solution that will cover several languages, however in this prototype implementation we focused on English texts only. 

In this paper, we present analysis and workflow inspired by the work of Joty, Carenini and Ng \cite{Joty2015}. We experimented with several methods in order to validate aspect-based sentiment analysis approaches and in the next steps we want to customise our implementation for the Polish language. 

The paper presents in Section \ref{sec:Introduction} an introduction to sentiment analysis and its importance in business, then in Section \ref{sec:RelatedWork} - related work from rhetorical and sentiment analysis areas is presented. Section \ref{sec:TheProposedMethod} covers description of our method. Implementation and the dataset are described in Section \ref{sec:ProblemDescription}. Section \ref{sec:Results} refers to the results. The last Section \ref{sec:Conclusions} consists of conclusions and future work.

\section{Related Work}
\label{sec:RelatedWork}

\subsection{Rhetorical Analysis}

\begin{figure}[!ht]
\centering
\includegraphics[scale=0.15]{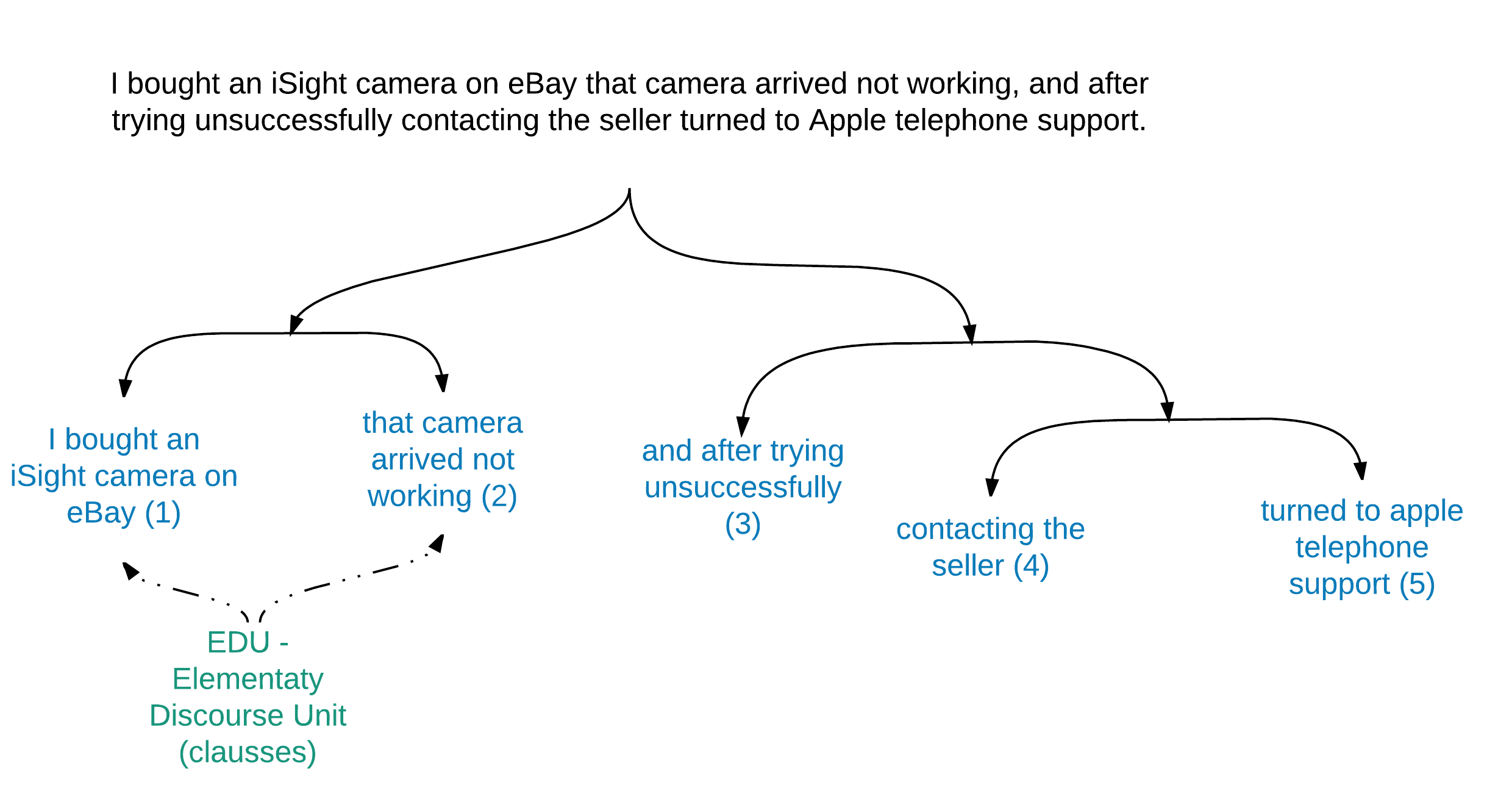}
\caption{An exemplary Discourse Tree based on Rhetorical Structure Theory.\label{fig:ra-example}}
\end{figure}

Rhetorical analysis seeks to uncover the coherence structure underneath the text, which has been shown to be beneficial for many Natural Language Processing (NLP) applications including text summarization and compression \cite{Louis2010}, machine translation evaluation \cite{Guzman2014}, sentiment analysis \cite{Lazaridou2013}, and others. Different formal theories of discourse analysis have been proposed. Martin \cite{Martin1992} proposed discourse relations based on discourse connectives (e.g., because, but) expressed in the text. Danlos \cite{Danlos2011} extended sentence grammar and formalize discourse structure. Rhetorical Structure Theory or RST - used in our experiments - was proposed by Mann and Thompson \cite{Mann1988}. The method proposed by them is perhaps the most influential theory of discourse in computational linguistics. Moreover, it was initially intended to be used in text generation tasks, but it became popular for parsing the structure of a~text \cite{Taboada2006}. Rhetorical Structure Theory represents texts by hierarchical structures with labels. This is a tree structure, which comprises Discourse Trees (DTs). Presented at Figure \ref{fig:ra-example} this Discourse Tree is a~representation of the following text:

%\begin{center}
%\textit{But he added: ''Some people use the purchasers' index as a~leading indicator, some use it as a~coincident indicator. But the thing it’s supposed to measure— manufacturing strength—it missed altogether last month.''}
% \end{center}

\subsection{Sentiment Analysis}

A sentiment analysis can be made at the level of (1) the whole document, (2) the individual sentences, or (what is currently seen as the most attractive approach) (3) at the level of individual fragments of text. Regarding document level analysis \cite{Augustyniak2015, Liu2010} - the task at this level is to classify whether a~full opinion expresses a~positive, negative or neutral attitude. For example, given a~product review, the model determines whether the text shows an overall positive, negative or neutral opinion about the product. The biggest disadvantage of document level analysis is an assumption that each document expresses views on a~single entity. Thus, it is not applicable to documents which evaluate or compare multiple objects. As for sentence level analysis \cite{Joty2012} - The task at this level relates to sentences and determines whether each sentence expressed a~positive, negative, or neutral opinion. This level of analysis is closely related to subjectivity classification which distinguishes sentences (called objective sentences) that express factual information from sentences (called subjective sentences) that express subjective views and opinions. However, we should note that subjectivity is not equivalent to sentiment as many objective sentences can imply opinions. With feature/aspect level analysis \cite{Wang2015} - both the document level and the sentence level analyses do not discover what exactly people liked and did not like. A~finer-grained analysis can be performed at aspect level. Aspect level was earlier called feature/aspect level. Instead of looking at language constructs (documents, paragraphs, sentences, clauses or phrases), aspect level directly looks at the opinion itself. It is based on the idea that an opinion consists of a~sentiment (positive or negative) and a~target (of opinion). As a~result, we can aggregate the opinions. For example, the phone display gathers positive feedback, but the battery is often rated negatively. The aspect-based level of analysis is much more complex since it requires more advanced knowledge representation than at the level of entire documents only. Also, the documents often consist of multiple sentences, so saying that the document is positive provides only partial information. In the literature, there exists some initial work related to aspects. There exist initial solutions that use SVM-based algorithms \cite{Wagner2014} or conditional random field classifiers \cite{DeClercq2015} with manually engineered features. There also exist some solutions based on deep neural networks, such as connecting sentiments with the corresponding aspects based on the constituency parse tree \cite{Wang2015}.

\section{Method for aspect-based sentiment analysis}
\label{sec:TheProposedMethod}

The proposed Rhetorical and Sentiment Analysis flow is divided into four main tasks: 
\begin{enumerate}
\item Rhetorical analysis with sentiment detection.
\item Aspect detection in textual data.
\item Methods, techniques, and graph analytics of aspect inter-relations.
\item Abstractive summary generation in natural language (not included in prototype workflow yet).
\end{enumerate}

\begin{figure}[!ht]
\centering
\includegraphics[scale=0.15]{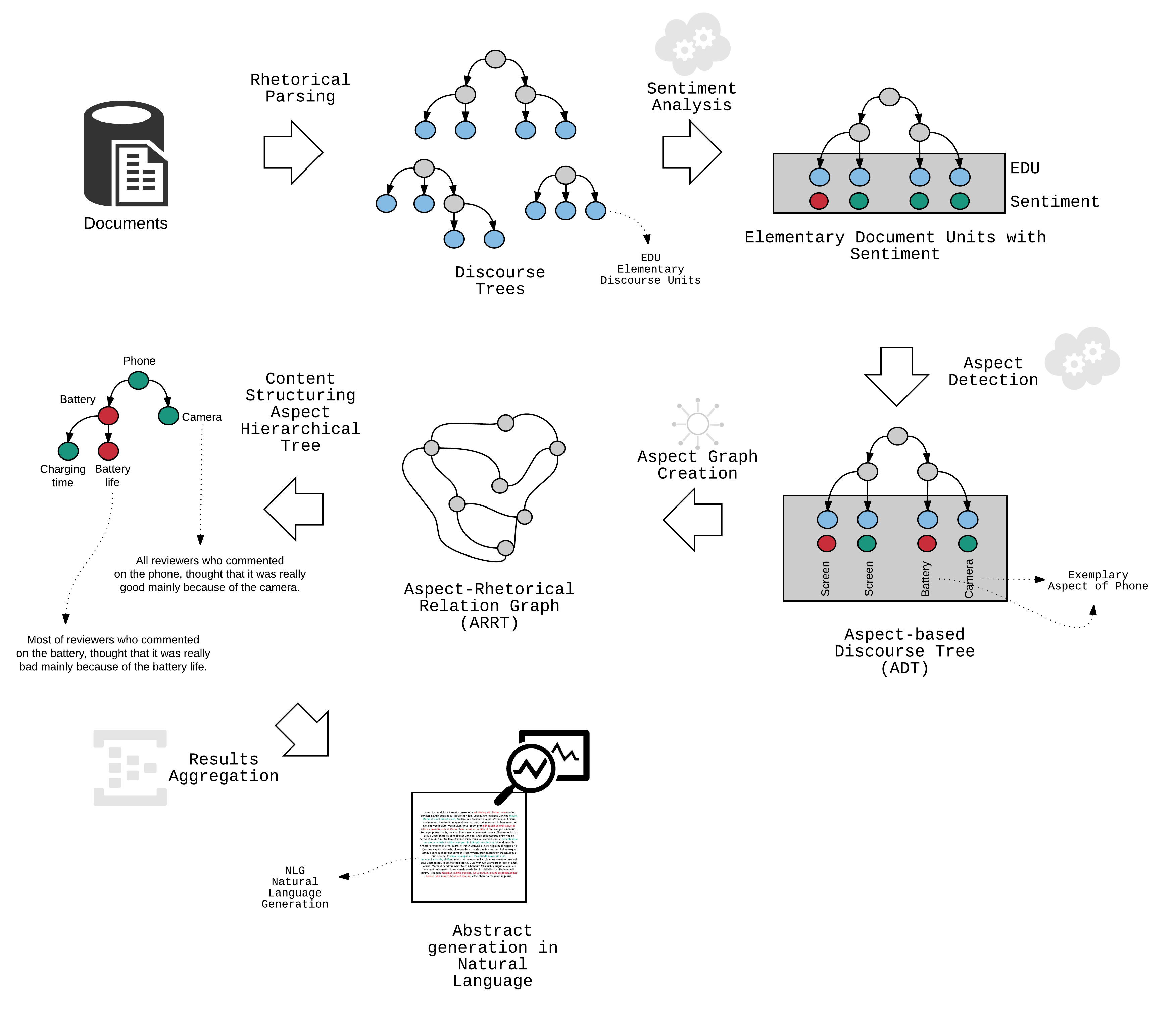}
\caption{The workflow for Rhetorical and Sentiment Analysis.\label{fig:flow}}
\end{figure}

The overall characteristics and flow organisation can be seen in Figure \ref{fig:flow}. Each of the mentioned steps of the proposed method is described in the following subsections.

\subsection{Rhetorical Analysis}

The goal of discourse analysis in our method is the segmentation of the text for the basic units of discourse structures EDU (Elementary Discourse Units) and connecting them to determine semantic relations. The analysis is performed separately for each source document, and as the output we get Discourse Trees (DT) such as in Figure \ref{fig:ra-example}. At this stage, existing discourse parsers will model the structure and the labels of a~DT separately. They do not take into account the sequential dependencies between the DT constituents. Then existing discourse parsers will apply greedy and sub-optimal parsing algorithms and build a~Discourse Tree. During this stage, and to cope with the mentioned limitation The inferred (posterior) probabilities can be used from CRF parsing models in a~probabilistic CKY-like bottom-up parsing algorithm \cite{Jurafsky2009} which is non-greedy and optimal. Finally, discourse parsers do not discriminate between intra-sentential parsing (i.e., building the DTs for individual sentences) and multi-sentential parsing (i.e., building a DT for the whole document) \cite{Joty2015}. Hence, this part of the analysis will extract for us distributed information about the relationship between different EDUs from parsed texts. Then we assign sentiment orientation to each EDU. 

\subsection{Aspect detection in textual data}
The second step covers aspect extraction and creation of aspect-based discourse trees ADT - see Figure \ref{fig:flow}. Aspect detection from textual data is based commonly on detection of names or noun-phrases \cite{MariaPontikiDimitriosGalanisHarisPapageorgiouSureshManandhar2015} and we used exactly this approach. 

% LDA (Latent Dirichlet Allocation) \cite{Jo2011}, lists of words that may be aspects \cite{Wilson2009}, detection of keywords and the most significant \cite{Mihalcea2004}, recently with wider and wider applicability and feasibility neural models, especially deep learning models has appeared \cite{Wang2015}.

\subsection{Analysis of aspect inter-relations}
The third step consists of an Aspect-Rhetorical Relation Graph (ARRG) and content Structuring Aspect Hierarchical Tree (see Figure \ref{fig:flow}). Discourse Trees of individual documents are processed (the order of EDU is not changed) to form association rules. Then, an Aspect-Rhetorical Relation Graph based on a~set of these rules is created. Each node represents an aspect and each edge is one of the relations between the EDU’s aspects. A graph will be created for all documents used in the experiment. The graph can be represented with weighted edges (association rules confidence, a number of such relations in the whole graph etc.), but there is a need to check and compare different types of graph representations. Then, it is possible to characterise the whole graph and each node (aspect) with graph metrics (PageRank \cite{Page1998}, degree, betweenness or other metrics). These metrics will be used for estimating the cut threshold – removing uninformative or redundant aspects. Hence, we will end up with only the most important aspects derived from analysed corpora. Then the graph will be transformed into an Aspect Hierarchical Tree. This represents the correlation between aspects and enables us to generate natural language-based descriptions.

\subsection{Abstractive summary generation in natural language}
The last step covers summary (abstract) generation in natural language. Natural language generation models use parameterized templates (very limited and dependent on the size of the rule-based system responsible for the completions of the text), or deep neural networks \cite{Wen2015}.

\section{Experimental Scenario}
\label{sec:ProblemDescription}

For the Rhetorical Parsing part of our experiment, we used a special library implemented for such purposes \cite{Feng2014}. As a sentiment analysis model, we used the Bag of Word vectorization method with a Logistic Regression classifier trained on 1.2 million (1, 3 and 5-star rating only) of Electronic reviews from SNAP Amazon Dataset \cite{McAuley2013}. The BoW vectorization method built a~vocabulary that considers the top 50,000 terms only ordered by their frequency across the corpus, similarly to supervised learning examples presented in our previous works in \cite{Augustyniak2015}. We used a noun and noun phrases extractor according to part-of-speech tagger from the Spacy Python library\footnote{\url{https://spacy.io}}. In order to create an Aspect-Rhetorical Relation Graph we used breadth-first search (BFS) algorithm for each Discourse Tree.

\subsection{Dataset}
\label{sec:dataset}

We used Bing Liu's dataset \cite{Liu2015} for evaluation. It contains three review datasets of three domains: computers, wireless routers, and speakers as in Table \ref{tab:dataset}. Aspects in these review datasets were annotated manually.

\begin{table*}[!h]
\caption{Bing Liu's dataset \cite{Liu2015} statistics.}
\label{tab:dataset}
\centering
\begin{tabular}{c|c|c}
Dataset & \# of documents & \# of distinct aspects \\
\hline
Computer & 531 & 354 \\
Wireless router & 879 & 307 \\
Speaker & 689 & 440 \\
\end{tabular}
\end{table*}

\subsection{Experimental Setup}

We implemented our framework in Python. The first computational step was to load the dataset and parse it into individual documents. Next, each document was processed through the Discourse Parser \cite{Feng2014} and transformed into a~Discourse Tree (DT). Then we extracted Elementary Discourse Units (EDUs) from the DT and each EDU was processed through the Logistic Regression sentiment algorithm. All neutral EDUs were taken off from consideration to ensure that the discovered aspects are correlated with authors' emotions. The remaining EDUs were processed through part-of-speech tagger to extract nouns and noun phrases which we decided to treat as potential aspects. The result of this step was a~set of Aspect-based Discourse Trees (ADTs). Then, from each ADT relations between aspects were extracted using breadth-first search, and an Aspect-Rhetorical Relation Graph (ARRG) was created by using aspects and relations such as nodes and edges respectively. Next, we evaluated the importance of aspects using a PageRank algorithm. Our approach resulted in complete list of aspects sorted by PageRank score. We applied a user-selected importance threshold to filter trivial aspects.

\section{Results}
\label{sec:Results}

In Table \ref{tab:results_compare} there are presented some examples of the results of our approach compared with the annotated data from Bing Liu's dataset. In the first sentence, the results of the analysis differ because we decided to treat only nouns or noun phrases as aspects, while annotators also accepted verbs. In some cases, such as sentences 2 or 4, our approach generated more valuable aspects than the annotators found, but in some cases, like sentence 5, we found fewer. This is possibly the result of our method of filtering valuable aspects - if some aspects were not frequent enough in the dataset, we can treat them as void. In cases where there is neither aspect nor sentiment in the dataset, such as sentence 6, we measure sentiment as well, as one of our analysis steps.

\begin{table}[ht]
% \hspace*{-1cm}\begin{tabular}.....\end{tabular}\hspace*{-1cm}
\centering
\caption{Examples of proposed analysis results}
% \hspace*{-2cm}
% \def\arraystretch{1.5}%  1 is the default, change whatever you need
\begin{tabular}{P{0.4cm} | P{6cm} | P{3.5cm} | P{3.3cm}}
 \textsc{No.} & \textsc{Input content} & \textsc{Annotated aspect : sentiment} & \textsc{Detected aspect : sentiment}  \\\hline
 1 & 
 I have this connected to my late 2008 MacBook Pro, and it works flawlessly. & 
 works : positive &
 macbook pro : positive \\\hline
 
 2 & 
 We are well pleased with the monitor and the company. & 
 monitor : positive &
 \shortstack{monitor : positive \\ company : positive} \\\hline

 3 & 
 The changing colors help to tell, with a~quick glance. & 
 colors : positive &
 colors : positive \\\hline
 
 4 & 
 The screen is a~very pleasing matte, and the colors are great. & 
 colors : positive &
 \shortstack{screen : negative \\ colors : positive} \\\hline
 
 5 &
 I would not recommend this or any Acer product to anyone except perhaps my ex. & 
 Acer product : negative &
 - : negative \\\hline
 
 6 &
 I purchased this as a~Christmas gift &
 - : - &
 - : negative \\\hline
 
%  7 &
%  The only hard button controls you get are brightness and contrast. &
%  - : - &
%  - : neutral \\\hline
 
%  8 & 
%  The screen quality is very high and the side view is very sharp and clear. &
%  \shortstack{screen quality : positive \\ side view : positive} &
%  view : positive \\
\end{tabular}
% \hspace*{-2cm}
\label{tab:results_compare}
\end{table}

\begin{figure}[ht]
\includegraphics[width=.49\linewidth]{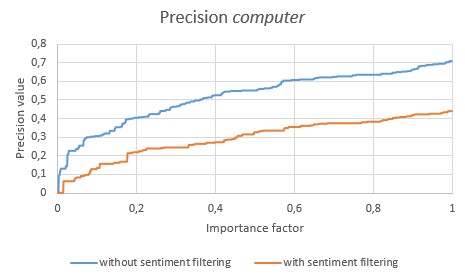}
\includegraphics[width=.49\linewidth]{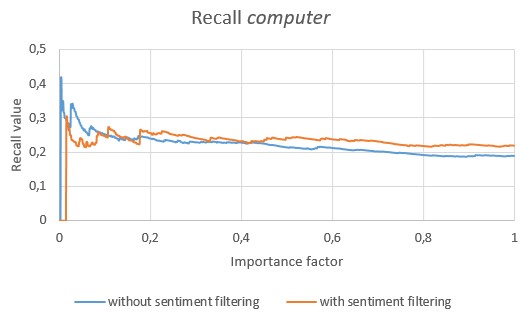}
\includegraphics[width=.49\linewidth]{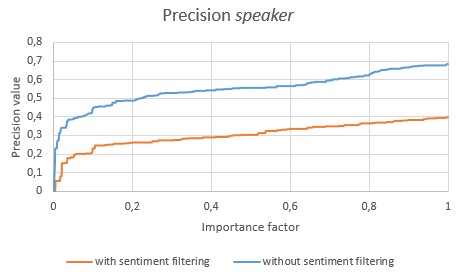}
\includegraphics[width=.49\linewidth]{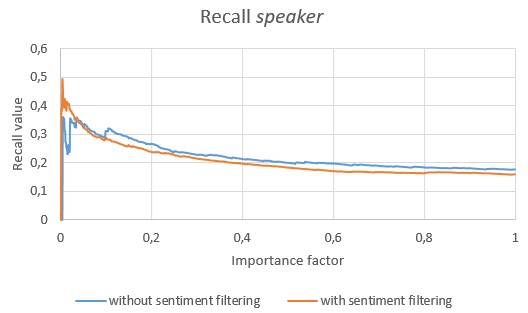}
\includegraphics[width=.49\linewidth]{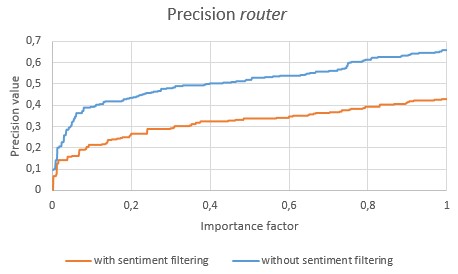}
\includegraphics[width=.49\linewidth]{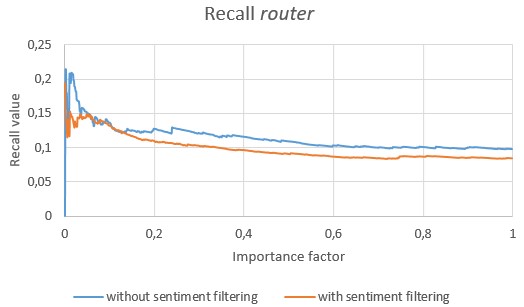}
\caption{Agreement between found aspects and gold standard with (blue line) and without sentiment filtering (orange line)}
\label{fig:results_prec_rec}
\end{figure}

Figure \ref{fig:results_prec_rec} shows the agreement between our aspects and that of the dataset. We assumed two aspects as equal when they were textually the same. We made some experiments using text distance metrics, such as the Jaro-Winkler distance, but the results did not differ significantly from an exact matching. We fitted the importance factor value (on the X axis) so as to enrich final aspects set: a higher factor resulted in a~larger aspects set and a~higher value of precision metric, with slowly decreasing recall. First results (blue line on charts) were not satisfactory, so we removed a~sentiment filtering step of analysis (orange line on chart), which doubled the precision value, with nearly the same value of recall. The level of precision for whole dataset (computer, router, and speaker) was most of the time at the same level. However, the recall of router was significantly worse than speaker and computer sets.  

\section{Conclusions and Future Work}
\label{sec:Conclusions}

We have proposed a~comprehensive flow of analysing aspects and assigning sentiment orientation to them. The advantages of such an analysis are that: it is a~grammatically-based and coherent solution, it shows opinion distribution, it doesn't need any aspect ontology, it is not limited to the number of aspects and really important, it doesn't need training data (unsupervised method). The method proved it has a big potential in generating summary overviews for aspect and sentiment distribution across analysed documents. In our next steps, we want to improve the aspect extraction phase, probably using neural network approaches. Moreover, we want to expand the analysis of the Polish language.

\section*{Acknowledgment}
The work was partially supported by the National Science Centre grants DEC-2016/21/N/ST6/02366 and DEC-2016/21/D/ST6/02948, and from the European Union’s Horizon 2020 research and innovation programme under the Marie Skłodowska-Curie grant agreement No 691152 (RENOIR project).

\bibliographystyle{plain}   % basic style, author-year citations
\bibliography{aciids}  % name your BibTeX data base
\end{document}